\pdfoutput=1
\documentclass[11pt]{article}

\usepackage[]{EMNLP2022}

\usepackage{times}
\usepackage{latexsym}

\usepackage{hyperref}

\usepackage[T1]{fontenc}
\usepackage[utf8]{inputenc}

\usepackage{microtype}

\usepackage{inconsolata}

\newcommand{\ChildTuningD}{{Child-Tuning}$_D$\xspace}

\newcommand{\ChildTuningF}{{Child-Tuning}$_F$\xspace}

\usepackage{xspace}
\usepackage{graphicx}
\usepackage{amsmath}
\usepackage{algorithm}
\usepackage{algorithmic}

\usepackage{mathtools}
\usepackage{amsfonts}

\title{HyPe: Better Pre-trained Language Model Fine-tuning with Hidden Representation Perturbation}

\author{Hongyi Yuan$^{12}$\thanks{$\quad$Work done at Alibaba DAMO Academy.}  , Zheng Yuan$^2$, Chuanqi Tan$^2$, Fei Huang$^2$, Songfang Huang$^2$   \\                         
  $^1$Tsinghua University, $^2$Alibaba Group \\
  \texttt{yuanhy20@mails.tsinghua.edu.cn} \\
  \texttt{\{yuanzheng.yuanzhen,chuanqi.tcq,f.huang,songfang.hsf\}@alibaba-inc.com}
  }
\begin{document}
\maketitle
\begin{abstract}
% Pre-trained language models (PLMs) have shown great performance fine-tuning on the downstream natural language processing tasks. By stacking Transformer layers, large PLMs can have millions of parameters which enable PLMs with great ability to fit data. However, PLMs may be provided limited data when fine-tuning, thus can easily over-fit. Perturbing data with noise is proven effective to alleviate over-fitting in machine learning literature. In this paper, we take the common stacking structure of PLMs into consideration and propose Layer-wise Noise. HyPe is an effective fine-tuning technique that mitigates over-fitting problem by perturbing the hidden representations between layers of PLMs with noise. HyPe improves PLM fine-tuning by stabilizing representations from every layer. Through extensive experiments, we empirically demonstrate that fine-tuning with HyPe outperforms vanilla fine-tuning and achieve better results than the previous state-of-the-art fine-tuning techniques. Further analysis also provides insights on how HyPe works.  

% Fine-tuning the Pre-trained language models (PLMs) is the main paradigm for advancing natural language processing methods.

% problems of unstable fine-tuning performances such as over-fitting and occasional training failure.
Language models with the Transformers structure have shown great performance in natural language processing.
However, there still poses problems when fine-tuning pre-trained language models on downstream tasks, such as over-fitting or representation collapse.
In this work, we propose HyPe, a simple yet effective fine-tuning technique to alleviate such problems by perturbing hidden representations of Transformers layers. Unlike previous works that only add noise to inputs or parameters, we argue that the hidden representations of Transformers layers convey more diverse and meaningful language information. 
Therefore, making the Transformers layers more robust to hidden representation perturbations can further benefit the fine-tuning of PLMs en bloc.
We conduct extensive experiments and analyses on GLUE and other natural language inference datasets. Results demonstrate that HyPe outperforms vanilla fine-tuning and enhances generalization of hidden representations from different layers. In addition, HyPe acquires negligible computational overheads, and is better than and compatible with previous state-of-the-art fine-tuning techniques. Codes are released at \url{https://github.com/Yuanhy1997/HyPe}.
% Therefore, perturbing hidden representations can make the Transformer layers more robust and further benefit the fine-tuning of PLMs en bloc. 

% Perturbing input features is an effective strategy to augment data thus alleviate over-fitting. Diverse language information is resolved by different Transformer layers and is encoded into hidden representations. Therefore, 
% in this paper, we propose HyPe, an effective fine-tuning technique that perturbs hidden representations in PLMs with random noise. By adding random noise to hidden representations, HyPe improves fine-tuning of Transformer layers hence fine-tuning of PLMs en bloc. 

\end{abstract}

\section{Introduction}

Pretrain-then-finetune has become the mainstream paradigm in recent natural language processing (NLP) practices, and there emerges various pre-trained language models (PLMs) such as BERT \cite{Devlin2019BERTPO}, RoBERTa \cite{Liu2019RoBERTaAR}, and XLNet \cite{xlnet}.
Vanilla PLM fine-tuning with common strategies (e.g., dropout \cite{dropout} and AdamW \cite{loshchilov2018decoupled}) can empower PLMs with excellent downstream performance. However, vanilla fine-tuned PLMs acquire performances with large variances on the downstream tasks \cite{empirical_finetuning_dodge}. Such unstable performances may results from over-fitting or representation collapse \cite{r3f}.  
These problems can be aggravated in low-resource scenarios \cite{Zhang2021RevisitingFB}.

\begin{figure}[t]
    \centering
    \includegraphics[width=0.48\textwidth]{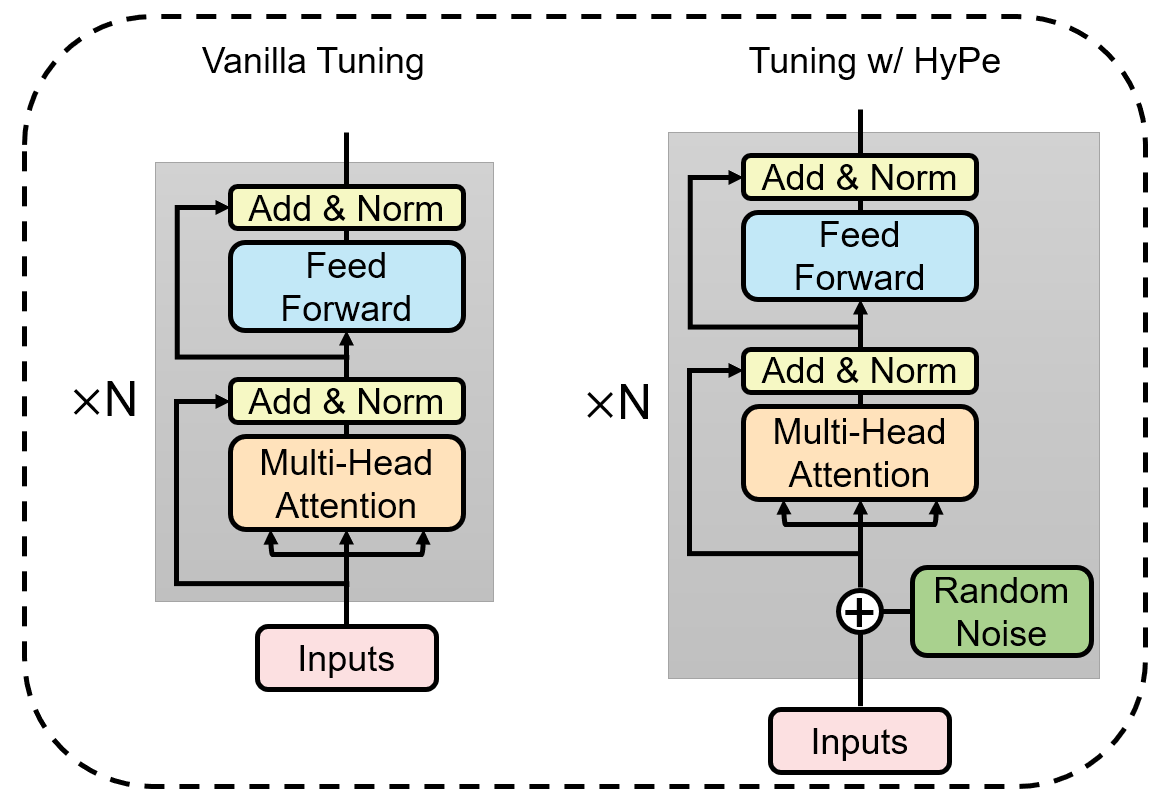}
    \caption{The overview of proposed HyPe fine-tuning technique. The random noise is added to the hidden representations fed into each Transformers layer in the forward computation of PLMs.}
    \label{fig:overview}
\end{figure}

In recent literature, effective fine-tuning techniques have been proposed to improve the performance and generalization (transferability) of fine-tuned PLMs \cite{jiang-etal-2020-smart, mixout,chen-etal-2020-recall}. 
Besides other explicit regularization, adding noise is a widely-used strategy to smoothen the optimization landscape and mitigate over-fitting.
For example, some works apply the perturbation to pre-trained parameter weights (e.g., NoisyTune \cite{wu-etal-2022-noisytune}), input embedding features (e.g., R3F \cite{r3f}) or gradients (e.g., ChildTuning \cite{xu-etal-2021-raise}) during the fine-tuning process.

Injecting noise to input features is a conventional technique for generalization and can be seen as implicit parameter regularization \cite{bishop}. Common PLMs are stacked basic neural network layers (i.e., Transformer layers \cite{transformers}), and previous research \cite{tenney-etal-2019-bert} points out that different Transformers layers of PLMs resolve different language information which is encoded in hidden representations. We turn to inject noise between layers to enhance the hidden semantic representations for better generalization on Transformers layer level.

Based on the above findings, we propose to improve fine-tuning by perturbing the hidden representations.
As shown in Figure \ref{fig:overview}, we propose a simple yet effective fine-tuning technique named \textbf{HyPe} (\textbf{Hi(y)}dden representation \textbf{Pe}rturbation) that adds random noise to the hidden representations between layers (i.e., the inputs of \textbf{each} Transformers layer) to alleviate the performance of fine-tuned layers from degrading.
To be concrete, we introduce no inductive biases to the distributions of noise in HyPe and focus on the pivotal influences of noise per se. Although noise can be compatible with auxiliary constrains \cite{r3f} or include informative priors \cite{xu-etal-2021-raise}, they may lead to non-negligible computational overheads. We simply use the uniform and normal distributions as two variants of noise distributions and denote them as HyPe-U and HyPe-N, respectively. The computational overheads are marginal in HyPe.
HyPe can also be regarded as a decoupling analysis of the above methods.

We conduct extensive experiments on GLUE benchmark \cite{wang-etal-2018-glue} and HyPe improves vanilla fine-tuning up to 1.60 on BERT in terms of average scores of the relatively small datasets MRPC, RTE, CoLA, and STS-B, surpasses previous state-of-the-art techniques (i.e. R-Drop \cite{liang2021rdrop}) by 0.15, and improves performance in low-resource scenarios. Further analyses demonstrate that HyPe
is also compatible with different scales of PLM (Section \ref{sec:scaling}) and other fine-tuning techniques (Section \ref{sec:compatible}), increases the robustness towards adversarial attacks (Section \ref{sec:advglue}), and improves generalization across tasks and domains on different layers (Section \ref{section:gen}).

To summarize our work, the main contributions are listed as follows:
\begin{enumerate}
    \item We propose HyPe, a simple yet effective fine-tuning technique requiring little computational overhead to improve the performance and transferability of fine-tuning PLMs. 
    \item Extensive experimental results show that 1) HyPe improves fine-tuning in the aspect of task performance and generalization and is complementary to PLM scaling; 2) HyPe surpasses and is compatible with current state-of-the-art fine-tuning techniques.
\end{enumerate}

\section{Related Works}

For large-scale PLMs, fine-tuning on downstream tasks may acquire unstable performances, resulting from over-fitting problems or failed training runs \cite{empirical_finetuning_dodge,Zhang2021RevisitingFB}. Recent research has focused on how to alleviate such problems and effectively improve fine-tuning of PLMs on the downstream tasks. 

A general idea is to make the best of the pre-trained weights and constrain the fine-tuned parameters from deviating much from the pre-trained weights. For example, Top-K Tuning \cite{topktuning} only fine-tunes the top-k layers of PLMs and keeps the lower pre-trained layers intact. Inspired by DropConnect \cite{DropConnect}, mixout \cite{mixout} randomly replaces the weights of parameters with their pre-trained values instead of zero. RecAdam \cite{chen-etal-2020-recall} introduces $L^2$ distance to penalize the change of weights from the pre-trained ones. ChildTuning \cite{xu-etal-2021-raise} applies task-free or task-driven masks on the gradients thus only a subset of parameters are changed during fine-tuning. SAGE \cite{liang2022no} uses differential updating step sizes for each parameter. Parameters with higher sensitivities are updated less aggressively where the computation of sensitivities is related to the pre-trained parameters in the cases of PLM fine-tuning.

Another line of work use noise to improve fine-tuning. R-Drop \cite{liang2021rdrop} uses KL divergence to regularize the discrepancy between the noised outputs produced by different dropout \cite{dropout} masks during fine-tuning. Recently proposed NoisyTune \cite{wu-etal-2022-noisytune} directly adds weight-aware noise to the pre-trained parameters before fine-tuning to improve performance. Based on the ideas of trust regions and adversarial training, FreeLB \cite{freelb}, SMART \cite{jiang-etal-2020-smart} and R3F \cite{r3f} are proposed to improve fine-tuning by introducing adversarial noise to the input representations during training. \citet{tong2022robust} create noised input representations by interpolating the representations between in-batch samples. The augmented fine-tuning data can alleviate over-fitting and help PLMs learn a smoother decision boundary.

Previous research has proven the pivotal role of noise in improving PLM fine-tuning. Our proposed technique looks into the PLMs and adds noise to the hidden representations. Previous works introduce regulations along with the added noise.
Generating random noise only requires little computational overheads, while additional regulations can cause non-negligible computational overheads in memory footprints or training time, such as R-Drop requiring two forward computations in each training step \cite{liang2021rdrop}, and \ChildTuningD \cite{xu-etal-2021-raise} requiring to pre-compute Fisher information matrices.

\section{Hidden Representation Perturbation}
\label{sec:method}

HyPe is motivated to improve fine-tuning of PLMs. Perturbing input features for better training performance is proven in effect in wide machine learning applications \cite{cv_noise,r3f}. The structure of PLMs is complicated and different layers may have diverse impacts on understanding languages \cite{tenney-etal-2019-bert}. Therefore, by perturbing the hidden representations, we can improve the performance of each layer hence the whole PLMs in fine-tuning processes. 
% The detailed presentation of HyPe is as follows.

In the vanilla fine-tuning setting of language models, we denote the mapping of a PLM comprising of $n$ network layers as $f_\theta(\cdot)$ and the classification head for the downstream task as $c_\psi(\cdot)$, where $\theta$ stands for the pre-trained parameters of the PLMs and $\psi$ represents the parameters of the classification head on top of the PLM. Here we have the whole forward mapping $\hat{y}=c_\psi(f_\theta(x))$, where $x$ and $\hat{y}$ are the embedded language inputs and predicted target labels respectively. The training objective is $\mathcal{L}(\theta, \psi) = \mathcal{L}\left(c_\psi(f_\theta(x)), y\right)$, where $\mathcal{L}$ is the loss function defined by tasks.

The basic layer block of nowadays PLMs (e.g., BERT) is Transformers \cite{transformers} which mainly comprises of multi-head self-attention mechanism and feed-forward neural network. By stacking the Transformers layers, the scales of PLMs can get larger (e.g., the base and large versions of BERT contain 12 and 24 layers respectively). Given the stacking structure of PLMs, $f_\theta(x)$ can be decomposed as:
\begin{align*}
f_\theta(x)& = g_{\theta^n}\circ g_{\theta^{n-1}}\circ \cdots g_{\theta^1}(x),
% & = g_{\theta^n}(g_{\theta^{n-1}}\cdots (g_{\theta^1}(x))),
\end{align*}
where $g_{\theta^i}(\cdot)$ is the mapping function of the $i$-th Transformers layer of the PLM, $\theta^i$ represents the parameters within layer $i$ and we have $\cup_{i=1}^n\theta^i = \theta$. Let $h^i$ represents the hidden states fed into the layer $i$, then $h^{i+1} = g_{\theta^i}(h^{i})$. As the input sequences may comprise multiple word tokens, without the loss of generality, we omit the token position and sample index marks for $x$, $y$ and $h^i$ for simplicity.   

During fine-tuning, HyPe injects parameter-independent noise to the hidden states (representations) of each layer, then for the $i$-th layer:
\begin{align*}
h^{i+1} &= g_{\theta^i}(h^{i} + \varepsilon^i)\\
&\coloneqq g^{\varepsilon^i}_{\theta^i}(h^{i}),
\end{align*}
therefore the whole feed-forward process of the PLM becomes:
\begin{align*}
f_\theta^{\text{HyPe}}(x)& = g^{\varepsilon^n}_{\theta^n}\circ g^{\varepsilon^i}_{\theta^{n-1}}\circ \cdots g^{\varepsilon^1}_{\theta^1}(x),
\end{align*}
where $\varepsilon^i$ is the random noise for layer $i$ and each entry is distributed as $\mathcal{N}(0,\sigma^2)$ or $\mathcal{U}(-\sigma, \sigma)$. With HyPe, the training objective is simply:
\begin{align*}
\mathcal{L}^{\text{HyPe}}(\theta, \psi) = \mathcal{L}\left(c_\psi(f^{\text{HyPe}}_\theta(x)), y\right).
\end{align*}
% where $\mathcal{L}(\cdot, \cdot)$ is the task objective in vanilla fine-tuning.

As shown above, HyPe is a simple and straightforward fine-tuning technique. It can be easily applied to different tasks and PLMs. 

\begin{algorithm}[t] 
\caption{Forward Propagation with \textcolor{blue}{HyPe} } 
\label{alg:Framwork} 
\begin{algorithmic}[1] 
\ENSURE Word Token Sequences $x$
\STATE $h^1 = \text{EmbeddingLayer}(x)$ 
\FOR{each $i$ in layer number $n$}
\STATE \textcolor{blue}{Generate $\varepsilon^i$ from $\mathcal{N}(0,\sigma^2)$ or $\mathcal{U}(-\sigma,\sigma)$,}
\STATE \textcolor{blue}{$h^i = h^i+\varepsilon^i$,\\
// $\triangleright$ Add Random Noise to Hidden States}
\STATE $h^{i+1} = g_{\theta^i}(h^i)$,
\ENDFOR
\STATE $\hat{y} = c_\psi(h^n)$.
\RETURN $\hat{y}$
\end{algorithmic}
\end{algorithm}

\section{Experiments}

\begin{table*}[t]
  \centering
  \resizebox{1.\textwidth}{!}{
  \begin{tabular}{l|ccccc|ccccc}
    Dataset & STS-B &COLA &MRPC &RTE & AVG & STS-B &CoLA &MRPC &RTE & AVG \\
    \hline
    &\multicolumn{5}{c|}{BERT}&\multicolumn{5}{c}{XLNet} \\
    Vanilla&90.07$_{0.67}$&63.63$_{1.82}$& 90.67$_{0.92}$&72.24$_{2.18}$&79.15&91.68$_{{0.06}}$&30.91$_{24.99}$&92.12$_{0.40}$&75.57$_{11.63}$&72.57 \\
    HyPe-N&\textbf{90.37}$_{0.43}$&\textbf{66.26}$_{1.90}$&91.98$_{1.11}$&74.37$_{1.64}$&\textbf{80.75}&91.87$_{{0.06}}$&\textbf{64.40}$_{\phantom{0}0.72}$&\textbf{92.66}$_{{0.12}}$&83.15$_{{\phantom{0}0.90}}$&\textbf{83.02}\\
    HyPe-U&90.31$_{{0.41}}$&65.48$_{{0.45}}$&\textbf{92.12}$_{{0.28}}$&\textbf{74.49}$_{{0.95}}$&80.60&\textbf{91.97}$_{0.10}$&58.05$_{\phantom{0}2.53}$&92.40$_{0.24}$&\textbf{83.27}$_{\phantom{0}1.04}$&81.42\\
    % \hline
    &\multicolumn{5}{c|}{RoBERTa}&\multicolumn{5}{c}{ELECTRA} \\
    Vanilla&91.90$_{0.11}$&65.55$_{{0.36}}$&92.09$_{{0.16}}$&81.71$_{2.13}$&82.81&92.27$_{{0.16}}$&46.41$_{32.83}$&93.49$_{0.86}$&88.33$_{{\phantom{0}0.45}}$ &80.13\\
    HyPe-N&92.22$_{0.12}$&\textbf{66.04}$_{1.83}$&92.04$_{0.58}$&82.79$_{1.51}$&83.27&\textbf{92.37}$_{0.06}$&\textbf{68.88}$_{{\phantom{0}0.98}}$&\textbf{94.00}$_{0.61}$&\textbf{88.45}$_{\phantom{0}1.56}$&\textbf{85.93}\\
    HyPe-U&\textbf{92.29}$_{{0.06}}$&65.77$_{1.22}$&\textbf{92.60}$_{0.71}$&\textbf{84.12}$_{{0.29}}$&\textbf{83.70}&92.20$_{0.16}$&51.01$_{25.34}$&93.91$_{{0.44}}$&\textbf{88.45}$_{\phantom{0}1.18}$&81.39\\
   
    \end{tabular}
  }
  \caption{Comparison results of HyPe and vanilla fine-tuning on relatively small datasets using different PLMs. The best results are in \textbf{bold}. The standard deviations for each results are shown in the subscripts. AVG means the average score of the four datasets. Vanilla fine-tuning on CoLA using XLNet and ELECTRA is highly unstable hence resulting in low average scores with high variances.}
  \label{table-main}
\end{table*}

\begin{table*}[t]
  \centering
  \small
%   \resizebox{1.\textwidth}{!}{
  \begin{tabular}{l|ccccc}
    Dataset &SST2&QNLI&QQP&MNLI&AVG \\
    \hline
    Vanilla&95.83$_{0.30}$&93.43$_{0.77}$&88.99$_{0.12}$&\textbf{90.58}$_{0.07}$&92.21\\
    HyPe-N&\textbf{96.06}$_{0.05}$&93.98$_{0.27}$&89.15$_{0.13}$&90.32$_{0.07}$&92.38\\
    HyPe-U&96.02$_{0.19}$&\textbf{94.19}$_{0.24}$&\textbf{89.25}$_{0.15}$&90.25$_{0.13}$&\textbf{92.43}\\
    \end{tabular}
%   }
  \caption{Comparison results of HyPe and vanilla fine-tuning on large GLUE datasets using RoBERTa. The best results are in \textbf{bold}. The standard deviations for each results are shown in the subscripts. AVG means the average score of the four datasets.}
  \label{table-large}
\end{table*}

In this section, we empirically demonstrate the effectiveness of HyPe through extensive experiments. We use GLUE benchmark \cite{wang-etal-2018-glue} to illustrate the performance of HyPe in comparison to vanilla fine-tuning.

\subsection{Datasets}
\paragraph{GLUE} GLUE is a widely-used benchmark designed for evaluating the natural language understanding abilities of models. Tasks in GLUE cover different aspects of language understanding including sentiment analysis, language acceptability, etc. Following \citet{xu-etal-2021-raise}, we mainly use four relatively small datasets STS-B \cite{stsb}, MRPC \cite{mrpc}, RTE \cite{rte} and CoLA \cite{cola}, as the over-fitting problem is more notable in the small data settings \cite{empirical_finetuning_dodge}. We also use other larger datasets SST2 \cite{sst2}, QNLI \cite{qnli}, QQP\footnote{https://quoradata.quora.com/First-Quora-Dataset-Release-Question-Pairs} and MNLI \cite{mnli} to further illustrate the performance of HyPe. 
We report performance on the development set since the test set labels are not released. The statistics of GLUE are listed in Appendix \ref{app:glue}.

\subsection{Experiment Settings}

% We conduct experiments using the PLM codes and chekpoints from Huggingface.
For all experiments listed in the following, we do grid search on the learning rates and report the average results over three different random seeds. 
We use the hidden representations of the first special token (e.g., [CLS] in BERT) for sentence representation. 
For our HyPe, we conduct experiments on two variants with different distributions of noise, denoted as HyPe-N where $\varepsilon\sim\mathcal{N}(0, \sigma^2)$ and HyPe-U where $\varepsilon\sim\mathcal{U}(-\sigma, \sigma)$. HyPe is only added during training. When using HyPe, we empirically find that turning off dropout will improve the technique's performance, which will be discussed in Section \ref{sec:dropout}. Therefore, we run experiments with HyPe using no dropout on hidden representations. For the more detailed settings concerning individual experiments, we list them in Appendix \ref{app:generalexp}$\sim$\ref{app:sim_cal}.

\subsection{Performance on GLUE}

To illustrate the generality of HyPe, we conduct experiments on the GLUE benchmark with four popular PLMs, BERT-large \cite{Devlin2019BERTPO}, RoBERTa-large \cite{Liu2019RoBERTaAR}, ELECTRA-large \cite{electra} and XLNet-large \cite{xlnet}. 
We use the PLMs from Huggingface Hub\footnote{For these four PLMs, we use \texttt{bert-large-cased}, \texttt{roberta-large}, \texttt{google/electra-large-discriminator}, and \texttt{xlnet-large-cased}, respectively.} \cite{wolf-etal-2020-transformers}.

We first evaluate HyPe on the four relatively small datasets from GLUE.
%  since fine-tuning PLMs is prone to over-fitting when training samples are limited. 
 As shown in Table \ref{table-main}, 
%  HyPe fine-tuning outperforms vanilla fine-tuning across different datasets and PLMs. 
 both variants of HyPe with different noise consistently improve the performance over vanilla fine-tuning. On average scores across tasks, the improvements are 1.60 on BERT, 0.89 on RoBERTa, 7.45 on XLNet, and 5.80 on ELECTRA, respectively.
 In addition, HyPe can help the model converge better on the CoLA dataset using XLNet and ELECTRA with smaller standard deviations.

We also evaluate HyPe on relatively large datasets. We fine-tune RoBERTa on the larger datasets of GLUE benchmark, with and without HyPe. The results listed in Table \ref{table-large} also show that HyPe improves performance with large amounts of fine-tuning samples. The average gains across datasets are 0.22 and 0.17 for HyPe-U and HyPe-N respectively. 

In summary of the aforementioned results, we can conclude that HyPe improves and stabilizes fine-tuning consistently across different datasets and PLMs. 
In addition, we observe that the improvements are more significant on small datasets, which indicates that HyPe has the capability of mitigating the over-fitting problem of PLM fine-tuning.

\subsection{Performance with Low Resources}
\label{sec:low-resource}

\begin{table}[t]
  \centering
%   \resizebox{1.\textwidth}{!}{
    \small
  \begin{tabular}{l|ccc}
    Dataset &Vanilla&HyPe-N&HyPe-U \\
    \hline
    % &\multicolumn{4}{c}{RoBERTa}\\
    STS-B&89.28$_{\phantom{0}0.07}$&89.33$_{0.59}$&\textbf{89.77$_{0.41}$} \\
    CoLA&43.20$_{12.26}$&55.34$_{1.70}$&\textbf{56.34$_{2.23}$} \\
    MRPC&88.02$_{\phantom{0}0.80}$&\textbf{89.74$_{1.48}$}&88.49$_{0.11}$\\
    RTE&61.61$_{\phantom{0}6.95}$&74.61$_{7.32}$&\textbf{78.58$_{5.02}$} \\
    SST2&92.47$_{\phantom{0}0.68}$&\textbf{92.97$_{1.12}$}&92.51$_{0.47}$\\
    QNLI&84.94$_{\phantom{0}1.14}$ &\textbf{85.39$_{1.61}$}&84.86$_{1.19}$\\
    QQP&73.92$_{\phantom{0}3.59}$&74.97$_{1.69}$&\textbf{76.38$_{0.77}$}\\
    MNLI&60.90$_{11.89}$&79.90$_{1.49}$&\textbf{80.17$_{0.73}$} \\
    MNLI-mm&62.56$_{11.43}$&80.97$_{1.49}$&\textbf{81.43$_{0.63}$}\\
    \hline
    AVG&72.99&80.36&\textbf{80.95}
    % \hline
    \end{tabular}
%   }
  \caption{Two variants of HyPe results in low-resource scenarios in comparison to vanilla fine-tuning. Best results are in \textbf{bold}, and standard deviations are marked in subscripts.}
  \label{table-lowres}
\end{table}

As the amount of training data becomes smaller, the over-fitting problem can be more severe. Since HyPe shows good performance in mitigating over-fitting on relatively small GLUE datasets, we create a low-resource setting to further illustrate the performance of HyPe. We follow previous research \cite{xu-etal-2021-raise} for the low-resource setting. In detail, we subsample the training samples of each dataset in GLUE benchmark to a training subset with 1k samples, and evaluate the performance using the original development set. 
% The experiments are conducted on RoBERTa-large.

As shown in Table \ref{table-lowres}, both variants of HyPe with RoBERTa-large outperform vanilla consistently. On average, the improvements brought by HyPe-N and HyPe-U are up to 7.37 and 7.96 respectively. On some datasets, the improvements are significant: for example, the improvements of HyPe-N and HyPe-U are up to 13.00 and 16.97 on RTE respectively.
% On the 1k training subset, HyPe can profoundly narrow down the gaps to the full data settings. 
In summary, HyPe can effectively prevent PLMs from over-fitting when fine-tuning in low-resource scenarios.

\begin{table}[t]
  \centering
  \small
  \resizebox{0.48\textwidth}{!}{
  \begin{tabular}{l|cccc|c}
    
     & STS-B  &COLA &MRPC &RTE & Avg. Imp.  \\
    \hline
    \textbf{Base}&91.58& 63.81&92.34&84.84&- \\
    /w HyPe&91.86& 65.08&93.07&85.44&+0.72 \\
    \hline
    \textbf{Large} &92.39&67.01&93.34&90.97&- \\
    /w HyPe&92.68&67.92&93.17&91.10&+0.29\\
    \hline
    \textbf{XL} &92.62&69.12&92.97&91.34&- \\
    /w HyPe&92.56&70.74&93.33&91.94&+0.63 \\
    \hline
    \textbf{XXL} &93.02&70.24&93.80&92.06&-\\
    /w HyPe&93.23&70.76&94.26&92.42&+0.39  \\
  \end{tabular}
  }
  \caption{Performances on models with different parameter sizes. Avg. Imp. represents Averaged Improvements.}
  \label{tab:scaling}
\end{table}

\begin{table*}[t]
  \centering
  \small
%   \resizebox{1.\textwidth}{!}{
  \begin{tabular}{l|cccc|c}
    Dataset & STS-B &COLA &MRPC &RTE&Average \\
    \hline
    Vanilla&90.07$_{0.67}$&63.63$_{1.82}$& 90.67$_{0.92}$&72.24$_{2.18}$&79.31 \\
    Top-K Tuning*&89.97&62.63&91.09&70.90&78.65 \\
    Mixout*&89.99&63.60&91.29&72.15&79.26\\
    RecAdam*&89.86&64.33&90.85&71.63&79.17\\
    LNSR* & 90.23 & 63.35 & 88.50 & 73.31 & 78.85\\
    \ChildTuningF&90.24$_{0.45}$&63.86$_{1.60}$& 91.43$_{1.11}$ & 73.77$_{2.09}$&79.83\\
    \ChildTuningD&\underline{90.34}$_{0.55}$& 64.48$_{1.29}$& 91.43$_{0.24}$ & 73.65$_{0.51}$&79.97\\
    R-Drop&90.29$_{0.37}$&65.06$_{0.35}$&91.84$_{0.54}$&\textbf{75.21}$_{0.90}$&\underline{80.60}\\
    R3F&90.21$_{0.54}$&64.90$_{1.50}$&\textbf{92.23}$_{0.67}$&\underline{74.73}$_{2.41}$&80.52\\
    NoisyTune&90.22$_{0.55}$&64.67$_{0.27}$&91.46$_{0.64}$&73.89$_{1.78}$&80.06 \\
    \hline
    HyPe-N&\textbf{90.37}$_{0.43}$&\textbf{66.26}$_{1.90}$&91.98$_{1.11}$&74.37$_{1.64}$&\textbf{80.75} \\
     HyPe-U&90.31$_{0.41}$&\underline{65.48}$_{0.45}$&\underline{92.12}$_{0.28}$&74.49$_{0.95}$&\underline{80.60} \\
\end{tabular}
%   }
  \caption{Results comparing to other effective fine-tuning techniques using BERT-large. Best results are \textbf{bold}, and the second best results are \underline{underlined}. Standard deviations are provided in subscripts. * are reported from \citet{xu-etal-2021-raise} as the experiment settings are similar.}
  \label{table-compare}
\end{table*}

\begin{figure}[t]
    \centering
    \includegraphics[width=0.45\textwidth]{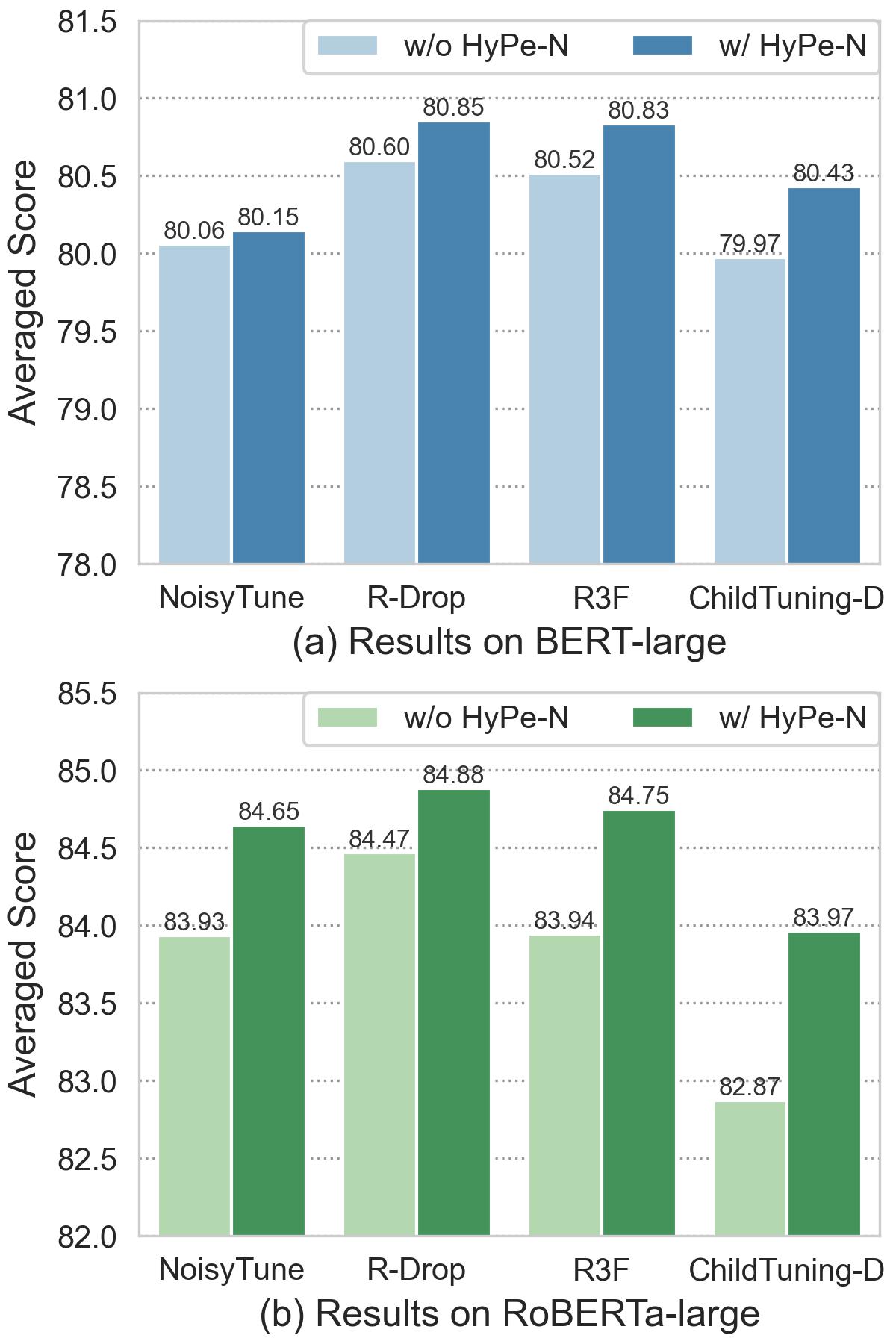}
    \caption{Results for combining HyPe-N with other effective fine-tuning techniques. The result numbers are the average evaluation scores of task MRPC, STS-B, CoLA, and RTE.}
    \label{fig:combine}
\end{figure}

\section{Further Analysis}

We provide further analyses and discussions on the performances of HyPe for model scaling, methods comparison and combination, adversarial attacks, and hyper-parameters in this section.

\subsection{Performance on Parameter Scaling}
\label{sec:scaling}

We investigate how HyPe performs as parameters of PLM scale up. We experiment on DeBERTa \cite{deberta} with 4 sizes: base, large, XL, and XXL. The experimental details are shown in Appendix \ref{app:scaling}. Results in Table \ref{tab:scaling} show that HyPe uniformly improves vanilla fine-tuning across different model sizes. The averaged improvements are +0.72, +0.29, +0.63, and +0.39 as the size scales up. This demonstrates that HyPe is complimentary to PLMs parameter scaling.

\subsection{Methods Comparison}

To compare HyPe with previous techniques for effective fine-tuning, we review and compare with the following baselines: (1) \textbf{Top-K Tuning} \cite{topktuning}; (2) \textbf{Mixout} \cite{mixout}; (3) \textbf{RecAdam} \cite{chen-etal-2020-recall}; (4) \textbf{R3F} \cite{r3f}; (5) \textbf{ChildTuning} \cite{xu-etal-2021-raise}; (6) \textbf{R-Drop} \cite{liang2021rdrop}; (7) \textbf{LNSR} \cite{lnsr}; (8) \textbf{NoisyTune} \cite{wu-etal-2022-noisytune}. The comparison experiments are conducted on the GLUE datasets STS-B, CoLA, MRPC, and RTE. 
% We fine-tune BERT for each dataset.

\paragraph{Comparison} From the results shown in Table \ref{table-compare}, HyPe achieves the best results on STS-B and CoLA, and consistently outperforms Top-K Tuning, Mixout, RecAdam, \ChildTuningF, and NoisyTune across different datasets. HyPe-N achieves the best average score of four tasks and surpasses the previous state-of-the-art R-Drop by 0.15. On MRPC and RTE, HyPe achieves competitive results with R3F, R-Drop, and \ChildTuningD. However, R3F and R-Drop include a KL divergence regularization objective and need to make two forward computations in a fine-tuning step. Both methods may have additional computational overhead. Take GPU memory footprints as an example, under the same training setting (e.g., batch size of 16), R3F and R-Drop require 16GB of memory while HyPe only requires about 11GB of memory. \ChildTuningD is a task-specific method and needs additional computation of the Fisher information matrix. HyPe only adds task-agnostic random noise to the hidden representations, and is more computationally efficient.

% \subsection{Combining with others}
\label{sec:compatible}

% In general, our HyPe is orthogonal to these methods. 
\paragraph{Compatibility} To show the complementarity of HyPe with other effective fine-tuning techniques, we conduct experiments on the combination of techniques. 
% In this section, we discuss whether there will be a further performance boost when combining HyPe with other effective fine-tuning techniques.
We integrate HyPe-N with four recently proposed state-of-the-art techniques, R-Drop, R3F, \ChildTuningD, and NoisyTune. We use MRPC, STS-B, CoLA, and RTE datasets and apply different combinations to RoBERTa and BERT.
The average results of the four tasks in Figure \ref{fig:combine} show that combining HyPe with other effective fine-tuning techniques can further boost performance. This illustrates that the improvements brought by adding noise to hidden representations do not overlap with other techniques, thus another advantage of HyPe is being compatible with others. 
The details of experiment settings and results are shown in Appendix \ref{app:baseline}.

\subsection{Performance on Adversarial Samples}
\label{sec:advglue}

\begin{table}[t]
  \centering
  \small
  \resizebox{0.48\textwidth}{!}{
  \begin{tabular}{l|ccccc}
     advGLUE & SST-2 & MNLI(m/mm) & RTE & QNLI & QQP \\
    \hline
    Vanilla &33.03 & 28.72/27.05 & 40.46 & 39.77 & 37.91 \\
    HyPe &\textbf{34.45} & \textbf{32.51/27.78} & \textbf{48.56} & \textbf{47.97} & \textbf{40.17} \\
    \end{tabular}
    }
  \caption{Accuracy results on the adversarial attacked testing samples from advGLUE using BERT-large. Detailed data introduction and experiment settings are in Appendix \ref{app:advglue}. MNLI(m/mm) stands for MNLI-match/mismatch.}
  \label{table-advglue}
\end{table}

Fine-tuning PLMs may prone to bad generalization of adversarial attacks. Results listed in Table \ref{table-advglue} on textually crafted adversarial samples from advGLUE \cite{advglue} show that vanilla fine-tuned PLMs suffer from adversarial attacks, and compared to vanilla, the performance gains brought by HyPeN are up to +1.42, +3.79/+0.73, +8.10, +8.20 and +2.26 on advSST-2, advMNLI(m/mm), advRTE, advQNLI and advQQP respectively. The results demonstrate that injecting noise into the hidden representations can increase the robustness of fine-tuning towards adversarial attacks.

\begin{table*}[t]
  \centering
%   \resizebox{1.\textwidth}{!}{
    \small
  \begin{tabular}{l|ccccc|ccccc}
    &\multicolumn{5}{c}{Fine-tune on MNLI} & \multicolumn{5}{c}{Fine-tune on SNLI}  \\
    &Vanilla&HyPe-N&$\Delta$&HyPe-U&$\Delta$&Vanilla&HyPe-N&$\Delta$&HyPe-U&$\Delta$\\
    \hline
    SNLI&90.67&91.30&\textbf{+0.63}&90.77&\textbf{+0.10}&92.99&93.60&\textbf{+0.61}&93.49&\textbf{+0.50}\\
    SICK&90.30&89.76&-0.54&89.16&-1.14&87.74&89.09&\textbf{+1.35}&90.30&\textbf{+2.56}\\
    SciTaiL&80.04&81.40&\textbf{+1.36}&80.44&\textbf{+0.40}&79.58&80.71&\textbf{+1.13}&80.83&\textbf{+1.25} \\
    QQP&75.84&76.22&\textbf{+0.38}&76.04&\textbf{+0.20}&74.12&75.12&\textbf{+1.00}&74.90&\textbf{+0.78} \\
    MNLI&89.91&90.42&\textbf{+0.51}&90.01&\textbf{+0.10}&86.66&87.63&\textbf{+0.97}&87.40&\textbf{+0.74}\\
    MNLI-mm&90.73&91.12&\textbf{+0.39}&90.82&\textbf{+0.09}&87.28&88.44&\textbf{+1.16}&88.03&\textbf{+0.75}\\
    \end{tabular}
%   }
  \caption{Comparison results on domain generalization. $\Delta$ represents the change of performance over vanilla fine-tuning. Improvements of two HyPe variants over vanilla fine-tuning are in \textbf{bold}. All evaluation datasets are out-of-domain except the dataset from the training set itself.}
  \label{table-domain}
\end{table*}

\subsection{Performance on Generalization}
\label{section:gen}

Probings on generalization abilities is another scope to access the over-fitting problem of fine-tuning \cite{xu-etal-2021-raise, r3f}. In this subsection, we discuss the transferability of HyPe fine-tuned PLMs from the perspective of task generalization and domain generalization.

% \begin{figure}[t]
% \includegraphics[width=0.48\textwidth]{token_similarity_ver2.png}
% \caption{Token representation cosine similarity within samples across different layers. We use the fine-tuned RoBERTa checkpoints for all the results.}
% \label{figure:sim}
% \end{figure}
\begin{figure*}[t]
\includegraphics[width=1\textwidth]{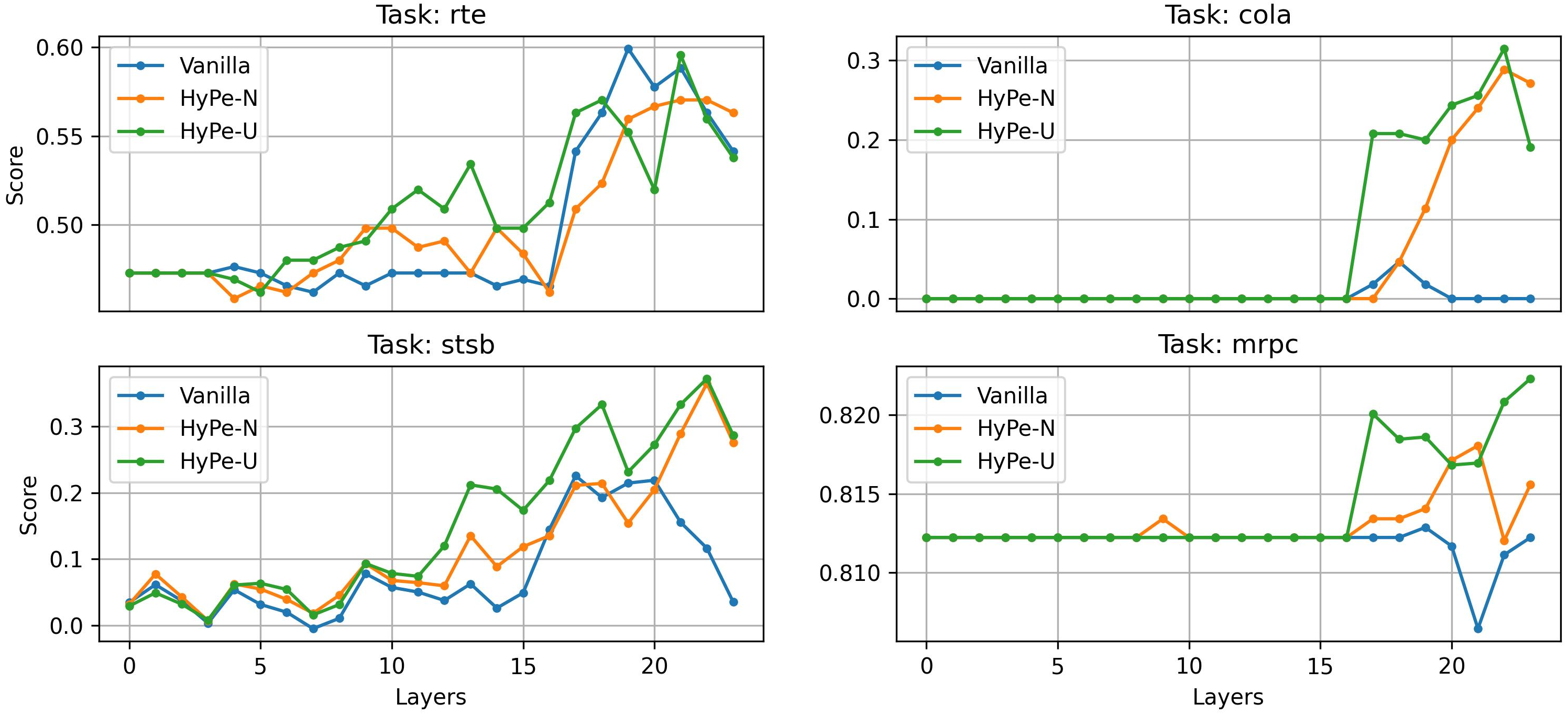}
\caption{Linear probings on the hidden representations across layers based on the RoBERTa-large checkpoint fine-tuned on SST2.}
\label{figue:task}
\end{figure*}

\paragraph{Task Generalization Probing}

One side effect of over-fitting is the degeneration of the dense representations of PLMs after fine-tuning, and the phenomenon is named representation collapse \cite{r3f}. We probe fine-tuned PLMs task generalization by training a PLM on one task and then evaluating on another with parameters fixed. Previous works freeze the whole parameters of PLMs and only tune a linear classifier for other tasks \cite{r3f,xu-etal-2021-raise}. As HyPe perturbs hidden representations among layers, we extend this experiment by training separated linear classifiers for hidden representation of each layer, and show their representational abilities.

We use MRPC, STS-B, RTE, and CoLA for the target tasks and start from the checkpoints of RoBERTa fine-tuned on SST2. As depicted in Figure \ref{figue:task}, it is shown that 1) both variants of HyPe achieve better performance than vanilla fine-tuning overall; 2) the improvement is more significant on higher layers of the PLM. In the lower layers, the three lines seem entangled. This is reasonable as the lower layers of PLMs are changed less in fine-tuning, as discussed by previous research \cite{durrani-etal-2021-transfer}. The results show that PLMs fine-tuned with HyPe maintain better representation ability across layers, thus demonstrating that they suffer less from the over-fitting problem.

\paragraph{Domain Generalization Probing} 
Besides generalization across tasks, \citet{xu-etal-2021-raise} also experiments on transferability across domains for the same. Good domain generalization may indicate that PLMs are fine-tuned to learn general semantic features and not easily over-fit the domain-specific information within training data. Following their work, we use natural language inference (NLI) tasks from different domains. Beyond NLI datasets MNLI and QQP in GLUE, we additionally introduce datasets SNLI \cite{snli}, SciTaiL \cite{Khot2018SciTaiLAT} and SICK \cite{sick}. For MNLI, we use both development sets of MNLI-match (MNLI) and MNLI-mismatch (MNLI-mm) for evaluation. Following previous research, we fine-tune RoBERTa-large with different techniques on a 5k sample subset of MNLI and SNLI datasets, respectively. Then, we test the fine-tuned PLMs on the aforementioned datasets to show the domain generalization ability. 
The detailed introductions of the datasets, experiment settings, and necessary label mappings are shown in Appendix \ref{app:gen}.

The results listed in Table \ref{table-domain} illustrate that both variants of HyPe outperform vanilla fine-tuned models on most of the out-of-domain datasets, except for SICK when fine-tuned on MNLI. 
This shows that HyPe can mitigate model over-fitting to domain-related features.
% by helping maintain the pre-training information and focusing on more general semantic features for the task when fine-tuning. 
Therefore when the domain of downstream tasks varies, PLMs fine-tuned with HyPe can still have good performance. 

Both generalization probing experiments above demonstrate that HyPe can help PLMs avoid representation collapse and over-fitting to the fine-tuning data, hence obtaining good generalization across tasks and domains.

\subsection{Discussions}
\label{sec:dropout}
\paragraph{Do the noise forms and scales matter?} Here we discuss how performance varies given different noise distributions and scales $\sigma$.

\begin{table}[t]
  \centering
  \small
%   \resizebox{0.48\textwidth}{!}{
  \begin{tabular}{l|cccc}
    Dataset & STS-B &CoLA &MRPC &RTE \\
    \hline
    Vanilla&90.07&63.63& 90.67&72.24\vspace{0.1cm}\\
    % \hline
    \multicolumn{5}{l}{HyPe-N}\\
    $\sigma=10^{-5}$&\textbf{90.37}&\textbf{66.26}&91.14&74.37 \\
    $\sigma=10^{-4}$&90.29&64.71&91.98&73.16\\ 
    $\sigma=10^{-3}$&\textbf{90.37}&64.94&91.73&72.80 \\ 
    $\sigma=10^{-2}$&90.36&64.60&91.61&74.13\vspace{0.1cm} \\
    % \hline
    \multicolumn{5}{l}{HyPe-U}\\
    $\sigma=10^{-5}$&90.24&65.48&\textbf{92.12}&73.65 \\
    $\sigma=10^{-4}$&90.31&65.13&91.83&\textbf{74.49} \\ 
    % \hline
    \end{tabular}
%   }
  \caption{Results of analysis experiments on the distribution forms and scales of the noise. Experiments conducted on BERT-large.}
  \label{table-ablate}
\end{table}

In Table \ref{table-ablate}, we can conclude from the results that 1) given different distributions and scales, HyPe consistently outperforms vanilla fine-tuning; 2) for different tasks the best choice for distributions and scales may differ: for example, on CoLA, the language acceptability task, the best choice is using a normal distribution with small scale $\sigma=10^{-5}$, while on MRPC, the semantical equivalence task, it is better to use uniform distribution with the scale of $\sigma=10^{-5}$.

% there exists no pattern of the impact of different scales when using the same distribution; 2) when the scale $\sigma$ is fixed, the two forms of distribution achieve similar performance; 3) all the results consistently outperform vanilla fine-tuning. Therefore, HyPe is insensitive to the choice of distributions and scales. It makes HyPe very easy to implement to be free from hyper-parameter searching.

\paragraph{Relation with Dropout} Note that in the aforementioned experiments we turn off dropout when using HyPe. When combining HyPe-N with dropout, we empirically find that the performance degrades. The average score drops from 80.75 to 79.92, as shown in Table \ref{table-dropout}. The possible explanation is that the improvement brought by dropout and that by HyPe partly overlap, since dropout randomly sets entries of hidden representations to zero, which can be regarded as a \textit{discrete} form of 0/1 noise \textit{multiplied} to different hidden representations where each entry of noise obeys a Bernoulli distribution. In terms of HyPe, we \textit{add} \textit{continuous} random noise to the hidden representations.  Empirically our HyPe shows superior performance than dropout, as in vanilla fine-tuning we apply 0.1 dropout rate. Therefore, adding continuous noise to the hidden representations in HyPe can be a good alternative for the discrete noise of dropout. 
% The explanation may be that when fine-tuning, dropout adds noise to the model parameters by randomly deactivating neurons under Bernoulli trails, while when inference, dropout is turned off. There, therefore, exists a discrepancy between training and inference which may hamper the downstream performance. HyPe can provide the noise needed for smoothening the optimization landscape and preventing over-fitting.

We leave the discussions of adding noise only to hidden representations of a subset of layers and adding additional noise to the representations of self-attention mechanism outputs inside each Transformers layer to Appendix \ref{app:more}.

\begin{table}[t]
  \centering
  \small
  \resizebox{0.49\textwidth}{!}{
  \begin{tabular}{l|ccccc}
    Dataset& STS-B &CoLA &MRPC &RTE&AVG \\
    \hline
    Vanilla&90.07&63.63& 90.67&72.24&79.15\\
    HyPe-N&\textbf{90.37}&\textbf{66.26}&\textbf{91.98}&\textbf{74.37}&\textbf{80.75}\\
    HyPe-N+DP&90.21&64.52&91.53&73.41&79.92 \\
    \end{tabular}
  }
  \caption{Results with and without dropout when using HyPe on BERT-large. DP represents dropout.}
  \label{table-dropout}
\end{table}

% \subsection{Summarization CNNDM}
% \subsection{Multi-modal Fine tuning}
% \subsection{Translation}

\section{Conclusion}

To conclude, we introduce HyPe, a technique to improve PLM fine-tuning. HyPe enhances fine-tuning by perturbing the intermediate hidden representations of a PLM with task and model agnostic random noise. Through experiments on GLUE and other NLI tasks, we demonstrate that PLMs fine-tuned with HyPe have better performance and transferability in comparison to vanilla fine-tuning, especially in a low-resource scenario. 
In further analyses, without additional regulation like KL-divergence and computational overheads, HyPe obtains superior performances compared to existing state-of-the-art fine-tuning techniques, and can further boost fine-tuning combined with others. 
Fine-tuning with HyPe improves hidden representations across different layers and provide stable improvements for generalization, adversarial attack and different model scales.

\section*{Limitations}

Collapsed fine-tuning runs mostly occur in the low resource scenario where PLMs may easily over-fit to the small data. The improvement with the proposed technique becomes marginal when the amount of training data scales up, as shown in Table~\ref{table-large}. The other limitation is that HyPe introduces two new hyper-parameters: The noise distribution form and the scale of variance. To achieve the best performance, we may need to search for different combinations of hyper-parameters.

\section*{Ethic Statement and Broader Impact}

As the parameter scale of PLMs and the pre-training cost get much larger hence showing better brilliant performance in language modeling, it is necessary to improve the fine-tuning performance of the language model in an effective and efficient way. Our proposed HyPe improves large PLM fine-tuning by only adding noise to the hidden representations. Unlike previous works, we do not include additional regulations since additional regulations may require non-negligible computational resources which may increase as the scale of PLM gets larger. It is important to develop effective fine-tuning techniques that are efficient and easy to implement. Through extensive discussions of HyPe, we illustrate that including perturbations in the features or representations could be the key part of why previous techniques work. Besides, we show that our HyPe can be a good continuous noise alternative for the widely-used dropout which can be regarded as 0/1 discrete noise multiplied to hidden representations. How and where to include perturbations and which forms of perturbations to apply to the fine-tuning of language models is worth studying and would be beneficial for advancing NLP frontiers.

\section*{Acknowledgments}
This work was supported by Alibaba Group through Alibaba Research Intern Program.

% We require little computational overheads, while additional regulation may need non-negligible computational resource which may increase as the scale of PLM gets larger. 

% Besides, we show that our HyPe can be a good continuous noise alternative for the widely-used dropout. Developing better training techniques that are more fitted to PLMs would be beneficial for advancing NLP frontiers. 

\bibliography{anthology,custom}
\bibliographystyle{acl_natbib}

\newpage

\appendix

\section{General Experiment Settings}
\label{app:generalexp}
On each experiment with each PLM, we run for three different random seeds for the averaged results and we grid search on learning rates of $\{1,2,3,4\}\times10^{-5}$ for the best results. Across different PLMs and tasks, we use AdamW \cite{loshchilov2018decoupled} as the optimizer with Adam $\beta$ of (0.9,0.99), Adam $\epsilon$ of $1\times10^{-5}$ and 0.1 weight decay. For the learning rate scheduler, we use a linear decay scheme. We truncate all the inputs to a length of 128 tokens. In vanilla fine-tuning, we use 0.1 dropout rate. For HyPe-N and HyPe-U, we use the best results of the scale $10^{-4}$ and $10^{-5}$ and turn off dropout if not otherwise specified.

All our experiments are conducted on 32G NVIDIA V100 GPU in a single GPU setting.

\section{Experiments on GLUE}
\label{app:glue}

\subsection{Data Introduction}

\begin{table}[th]
  \centering
  \resizebox{0.48\textwidth}{!}{
  \begin{tabular}{l|ccc}
    Dataset &Train. Size& Dev. Size & Metric  \\
    \hline
    MRPC&3.7k&408&F1 \\
    RTE& 2.5k&277&Accuracy\\
    STS-B&5.7k&1.5k&Pearson-Spearman Corr \\
    CoLA&8.5k&1.0k&Matthew's Corr \\
    QNLI&108k&5.7k&Accuracy \\
    QQP&364k&40k&F1 \\
    SST2&67k&872&Accuracy \\
    MNLI&393k&9.8k&Accuracy \\
    MNLI-mm&-&9.8k&Accuracy \\
    \end{tabular}  }
  \caption{The summary statistics of GLUE benchmark.}
  \label{tab:glue_sum}
\end{table}

The summary statistics of GLUE and the reported evaluation metric is listed in Table \ref{tab:glue_sum}. The license for GLUE is CC-BY-4.0.

%  5.7k/1.5k & 8.5k/1.0k & 3.7k/408 &  2.5k/277&67k/872&108k/5.7k&364k/40k&393k/20k

\subsection{Experiment Settings}

For different fine-tuning techniques, we experiment with the same hyper-parameter setting, which are listed in Table \ref{tab:glue_setting}.

\begin{table*}[t]
  \centering
 \small
  \begin{tabular}{lccccccc}
    Dataset &Batch Size&Update Steps&Warm-up Steps  \\
    \hline
    \textbf{BERT} \\
    MRPC&16&3 epochs&10\% of total steps  \\
    RTE&16&3 epochs&10\% of total steps \\
    STS-B&16&3 epochs&10\% of total steps \\
    CoLA&16&3 epochs&10\% of total steps  \\
    \hline
     \textbf{RoBERTa} \\
    MRPC&16&3 epochs&10\% of total steps \\
    RTE&16&3 epochs&10\% of total steps \\
    STS-B&16&3 epochs&10\% of total steps \\
    CoLA&16&3 epochs&10\% of total steps  \\
    SST2&16&3 epochs&10\% of total steps \\
    QNLI&16&3 epochs&10\% of total steps\\
    QQP&16&3 epochs&10\% of total steps\\
    MNLI&16&3 epochs&10\% of total steps\\
    \hline
     \textbf{ELECTRA} \\
    MRPC&32&3 epochs&10\% of total steps \\
    RTE&32&10 epochs&10\% of total steps\\
    STS-B&32&10 epochs&10\% of total steps\\
    CoLA&32&3 epochs&10\% of total steps \\
    \hline
     \textbf{XLNet} \\
    MRPC&32&800 steps&200 steps \\
    RTE&32&800 steps&200 steps\\
    STS-B&32&3000 steps&500 steps\\
    CoLA&64&1200 steps&120 steps\\
    \hline
    \end{tabular}
  
  \caption{Experiment settings used for different GLUE datasets and PLMs.}
  \label{tab:glue_setting}
\end{table*}

\subsection{GLUE Test Set Results}
\label{app:glue_test}

\begin{table*}[t]
  \centering
 \small
 \resizebox{1.0\textwidth}{!}{
  \begin{tabular}{lcccccccccc}
     &CoLA&STS-B&MRPC&RTE&AVG.($\Delta$)&SST-2&QNLI&QQP&MNLI-m/mm&AVG.-ALL($\Delta$)  \\
     
    \hline
    % \multicolumn{11}{l}{\textbf{RoBERTa}}\\
    Vanilla &62.3&90.7&90.8&79.9&80.93(-)&96.6&91.9&73.3&89.6/89.3&84.94(-)\\
    HyPe-N &65.5&90.9&91.0&81.0&82.10(+1.17)&96.5&94.1&73.0&89.8/89.6&85.71(+0.77)\\
    HyPe-U &65.2&91.1&92.3&82.8&82.85(+1.92)&96.4&93.8&73.1&89.9/89.6&86.02(+1.08)\\

    \hline
    
    \end{tabular}
  }
  \caption{Test set results on GLUE for RoBERTa-large. We use $\sigma=10^{-5}$ for HyPe-N and HyPe-U.}
  \label{tab:glue_test_results}
\end{table*}

The conventional evaluation procedures of the previous research (R3F, RDrop, ChildTuning, NoisyTune) only report results on development sets of GLUE. Here we compare the vanilla fine-tuned results with HyPe fine-tuned results on test sets. Results listed in Table \ref{tab:glue_test_results} show that on the averaged scores (column AVG.-ALL) of 8 GLUE tasks except WNLI and AX, HyPe-N and HyPe-U achieve 82.27 and 82.20 for BERT, as well as 85.71 and 86.02 for RoBERTa, which is obviously better than vanilla fine-tuning of 81.40 for BERT and 84.94 for RoBERTa. The improvements are more higher on 4 relatively small datasets (column AVG.), and HyPe-(N/U) achieves 2.30/1.90 and 1.17/1.92 for BERT and RoBERTa respectively. The results are consistent with those in Table \ref{table-main} and \ref{table-large} where HyPe can bring more performance gains on small data setting, since PLMs are prone to over-fitting more given small data.

\section{Generalization Probings}
\label{app:gen}

\subsection{Dataset Introduction}

The summary statistics of the NLI datasets SNLI, SICK and ScitaiL used in domain generalization probing experiments are presented in Table \ref{tab:nli_sum}. The licenses for SICK and ScitaiL are CC-BY-NC-SA-3.0 and  Apache-2.0 respectively.

\begin{table}[t]
  \centering
  \resizebox{0.48\textwidth}{!}{
  \begin{tabular}{l|cccc}
    Dataset &Train. Size& Dev. Size&Test Size & Metric  \\
    \hline
    SNLI&550,152&10,000&10,000&Accuracy \\
    ScitaiL&23,596&1,304&2,126&Accuracy \\
    SICK&4,439& 495&4,906&Accuracy \\
    \end{tabular}
  }
  \caption{The summary statistics of NLI datasets used in domain generalization probing experiments.}
  \label{tab:nli_sum}
\end{table}

\subsection{Experiment Settings}

\paragraph{Task Generalization} We freeze the model parameters fine-tuned on SST2 except for fine-tuning a re-initialized linear head for each task. For each experiment, we use a learning rate of 0.001 for 3 epochs and batch size 16 for tuning the linear heads.

\begin{table}[t]
  \centering
  \small
%   \resizebox{0.48\textwidth}{!}{
  \begin{tabular}{l|c}
  Dataset& Label Space\\
  \hline
  MNLI& entailment/neutral/contradiction\\
  MNLI-mm & entailment/neutral/contradiction \\
  SNLI& entailment/neutral/contradiction \\
  SciTaiL& entailment/neutral\\
  SICK& entailment/neutral/contradiction \\
  QQP& duplicate/not duplicate \\
 \end{tabular}
%  }
  \caption{The label spaces for datasets used in domain generalization experiments of Section \ref{section:gen}.}
  \label{tab:label_space}
\end{table}
\paragraph{Domain Generalization} We train on the subsets for 3 epochs with batch size 16. For different datasets we used, their label spaces are different as shown in Table \ref{tab:label_space}. Therefore, we follow the experiment settings in \citet{xu-etal-2021-raise}. Since SciTaiL only contains two labels \texttt{entailment} and \texttt{neutral} in their label spaces, we map the \texttt{contradiction} label in MNLI, MNLI-mm, SICK and SNLI to \texttt{neutral} to reduce their label space to \texttt{entailment} and \texttt{neutral}. For QQP, following \citet{gong2018natural}, we map \texttt{duplicate} to \texttt{entailment} and \texttt{not duplicate} to \texttt{contradiction}. With the above procedures, we create a consistent label space for each dataset to run evaluations. Besides, for some samples in SNLI, there exists no golden labels, and we filter them for training and evaluation. For the datasets used, we use their corresponding development sets for evaluation.

\section{With Other Techniques}
\label{app:baseline}

\subsection{Baseline Techniques}

Different previously proposed effective fine-tuning techniques have exclusive hyper-parameters, we list the hyper-parameters we used in our re-implementation in Table \ref{tab:baseline_settings}. For each, we follow the best settings reported in their papers. For ChildTuning, we use the Python code implementation from \url{https://github.com/alibaba/AliceMind/tree/main/ChildTuning}. For R-Drop, we use the implementation in \url{https://github.com/dropreg/R-Drop}. For R3F, we use the implementation from \url{https://github.com/facebookresearch/fairseq/tree/main/examples/rxf}. Note that in the original R3F implementation, they leave out STS-B task as this is a regression task and is not compatible with KL divergence. In our implementation, for STS-B task, we use mean squared error (MSE) in place of KL divergence for regulation.

\begin{table*}[t]
  \centering
  \small
%   \resizebox{0.48\textwidth}{!}{
  \begin{tabular}{lcc}
    \textbf{Technique} &\textbf{Hyper-parameters}&\textbf{Values} \\
    \hline
    \ChildTuningF&Gradient Mask Probability $p$&$\{0.2,0.3,0.4\}$ \\
    \ChildTuningD&Gradient Mask Probability $p$&$\{0.1,0.2,0.3\}$ \\
    R-Drop&Regularization Weight $\alpha$&$\{0.1,0.5,1.0\}$\\
    R3F&Noise Distribution& $\mathcal{N}(0,\sigma^2)$ \\
    &Noise Scale $\sigma$&$10^{-5}$ \\
    &Regularization Weight $\lambda$&$\{0.1,0.5,1.0\}$\\
    NoisyTune&Noisy Intensity $\lambda$&$\{0.1,0.15,0.2\}$ \\
    \end{tabular}
%   }
  \caption{The exclusive hyper-parameter settings for each baselines. For multiple values, we use the best results searched on these numbers.}
  \label{tab:baseline_settings}
\end{table*}

\subsection{Combination Experiments}
\label{app:how_combine}

We use the HyPe variant HyPe-N with scale $\sigma=10^{-5}$ to integrate with others.
When combining with \ChildTuningD, we add HyPe to the forward computations. When combining with R3F, we use HyPe for the noised forward computation. When combining with R-Drop, we add HyPe to two forward computations in a training step with no dropout. When combining with NoisyTune, we add the noise to the parameters before fine-tuning with HyPe. For the combination experiments, we also search on the same ranges of hyper-parameters for the best result.

\subsection{Detailed Results for Technique Combination}
\label{app:combine}
The detailed results for Figure \ref{fig:combine} are listed in Table \ref{tab:detail_combine}.

\begin{table*}[ht]
  \centering
  \small
%   \resizebox{1.\textwidth}{!}{
  \begin{tabular}{l|cccccc}
    Dataset & STS-B &COLA &MRPC &RTE&average&$\Delta$ \\
    \hline
    \textbf{Detailed results on BERT} \\
    RDrop&90.29$_{0.37}$&65.06$_{0.35}$&91.84$_{0.54}$&75.21$_{0.90}$&80.60&- \\
    HyPe-N+RDrop&90.45$_{0.33}$&65.23$_{0.43}$&91.80$_{0.26}$&75.93$_{0.85}$&80.85&+0.25\\ 
    R3F&90.21$_{0.56}$&64.90$_{1.50}$&92.23$_{0.67}$&74.73$_{2.41}$&80.52&-\\
    HyPe-N+R3F&90.36$_{0.37}$&65.58$_{0.52}$&91.82$_{0.44}$&75.57$_{0.85}$&80.83&+0.31\\
    \ChildTuningD&90.34$_{0.55}$& 64.48$_{1.29}$& 91.43$_{0.24}$ & 73.65$_{0.51}$&79.97&- \\
    HyPe-N+\ChildTuningD&90.75$_{0.65}$&65.18$_{1.17}$&91.77$_{0.30}$&74.01$_{0.29}$&80.43&+0.46\\
    NoisyTune&90.22$_{0.55}$&64.67$_{0.27}$&91.46$_{0.64}$&73.89$_{1.78}$&80.06&- \\
    HyPe-N+NoisyTune&90.37$_{0.51}$&65.12$_{2.12}$&91.45$_{0.20}$&73.65$_{0.29}$&80.15&+0.09 \\
    \hline
    \textbf{Detailed results on RoBERTa} \\
    RDrop&92.26$_{0.12}$&67.03$_{0.42}$&93.03$_{0.64}$&85.56$_{0.59}$&84.47&-\\
    HyPe-N+RDrop&92.34$_{0.03}$&68.77$_{3.59}$&93.21$_{0.90}$&85.20$_{2.36}$&84.88&+0.41\\ 
    R3F&92.13$_{0.08}$&67.32$_{1.72}$&92.32$_{0.68}$&84.00$_{1.62}$&83.94&-\\
    HyPe-N+R3F&92.29$_{0.07}$&68.25$_{0.42}$&92.64$_{0.72}$&85.80$_{1.70}$&84.75&+0.81\\
    \ChildTuningD&91.95$_{0.15}$&63.66$_{0.71}$&92.01$_{0.77}$&83.87$_{3.97}$&82.87&-\\
    HyPe-N+\ChildTuningD&92.05$_{0.28}$&67.38$_{1.35}$&92.31$_{0.37}$&84.12$_{0.51}$&83.97&+1.10 \\
    % \ChildTuningF&91.98$_{0.23}$&66.04$_{1.52}$&92.17$_{0.61}$&83.51$_{0.61}$ \\
    % HyPe-N+\ChildTuningF&92.19$_{0.12}$&68.06$_{1.23}$&92.63$_{0.31}$&84.24$_{1.04}$\\
    NoisyTune&92.07$_{0.21}$&66.15$_{0.13}$&92.31$_{1.02}$&85.20$_{0.59}$&83.93&- \\
    HyPe-N+NoisyTune&92.34$_{0.12}$&67.71$_{0.83}$&93.09$_{0.09}$&85.44$_{0.95}$&84.65&+0.72 \\
    \end{tabular}
%   }
  \caption{Detailed results of HyPe-N combining with other effective fine-tuning techniques. The standard deviations are shown in the subscripts.}
  \label{tab:detail_combine}
\end{table*}

\section{Experiment Details for advGLUE}
\label{app:advglue}

AdvGLUE \cite{advglue} contains the five adversarial perturbed datasets in GLUE which are SST-2, QQP, MNLI, RTE and QNLI. For MNLI there are MNLI-match and MNLI-mismatch. They use the original training data from the corresponding datasets in GLUE for model training. In our experiments, each results listed in Table \ref{table-advglue} are averaged out of 3 random seed runs.

\section{Experiment Details for Parameter Scaling Experiments}
\label{app:scaling}

When using vanilla fine-tuning schemes as settings listed in Table \ref{tab:glue_setting} will lead to corrupted and sub-optimal performances for DeBERTa. To reproduce a strong vanilla baseline for solid comparison, (1) we extend the training epochs to 6 and use a fixed warm-up step 100; (2) for MRPC, RTE and STS-B, we fine-tune based on MNLI-tuned models, which are \texttt{deberta-base-mnli}, \texttt{deberta-large-mnli}, \texttt{deberta-v2-xlarge-mnli} and \texttt{deberta-v2-xxlarge-mnli} from Huggingface repository, and for CoLA, we use the origin pre-trained versions , which are \texttt{deberta-base}, \texttt{deberta-large}, \texttt{deberta-v2-xlarge} and \texttt{deberta-v2-xxlarge} from Huggingface repository; (3) for the xlarge and xxlarge versions of DeBERTa's, we additionally search for best results on learning rates $\{1\times 10^{-6}, 3\times 10^{-6}, 5\times 10^{-6}, 8\times 10^{-6}\}$.

% \end{align*}
\section{More Discussions}
\label{app:more}

\subsection{Token Representation Similarity}
\label{app:sim_cal}

% \subsection{Token Representation Similarity}
% \label{sec:token-sim}
\begin{figure}[t]
\includegraphics[width=0.48\textwidth]{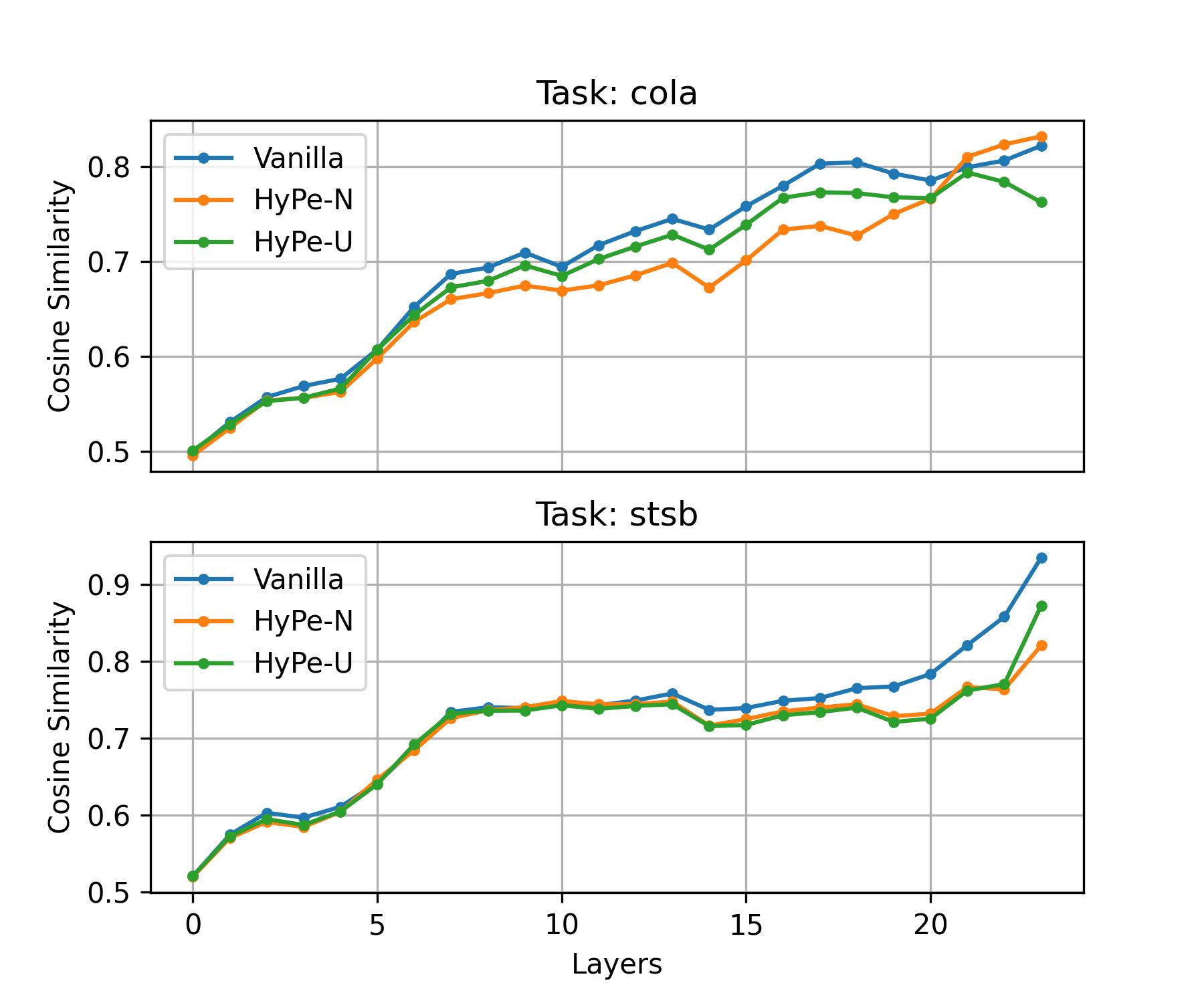}
\caption{Token representation cosine similarity within samples across different layers. We use the fine-tuned RoBERTa checkpoints for all the results.}
\label{figure:sim}
\end{figure}
As mentioned above in the generalization probing experiments, the representation abilities of hidden states are ameliorated. To further investigate how HyPe improves PLMs fine-tuning, we investigate the change of hidden representations. As illustrated by previous research \cite{ethayarajh-2019-contextual,gao2018representation}, PLMs may suffer from the problem of anisotropic distribution of token representations (i.e., the representations only distributed in a narrow cone of the entire high-dimensional space). Research finds a correlation between isotropic distribution of representations and downstream performance \cite{tacl,yu-etal-2022-rare}. Isotropic-distributed hidden representation is a good property in terms of good representation abilities. Representation anisotropy can be accessed by calculating the token-wise cosine similarity within a sample. The lower similarity indicates a more isotropic distribution. 

For the calculation of layer-wise token cosine similarity, we denote the index of each sample as $i$, the token index in each sample as $j$. The layer index is denoted as $l$. The calculation of similarity score $\mathcal{S}^l_i$ for layer $l$ and sample $i$ is:
\begin{align*}
    \mathcal{S}^l_i=\frac{2}{n_i(n_i-1)}\sum_{1\leq j_a < j_b \leq n_i}\cos(h^l_{ij_a}, h^l_{ij_b}),
\end{align*}
where $n_i$ is the token count of sample $i$, $h^l_{ij}$ stands for the hidden representation of token $j$ in sample $i$ in layer $l$ and $\cos$ stands for the cosine similarity $\cos(q,p) = \frac{q^Tp}{\|q\|\|p\|}$. Then the score is averaged over different samples:
\begin{align*}
    \mathcal{S}^l = \frac{1}{M}\sum_{i=1}^M \mathcal{S}^l_i,
\end{align*}
where $M$ is the number of samples.

With isotropic distribution where similarity values are larger, transformers layers do not show degeneration and maintain good representation capacities. Hidden states may carry diverse useful information to each token in the next layer throught attention mechanism. We investigate the similarity to provide insight on how HyPe improve final results.

In Figure \ref{figure:sim}, we provide a line plot on how hidden presentation similarity varies across layers. For each point, the results are averaged across samples and 3 different runs. We can see that the anisotropic distribution problem gets severe for the higher layers. Models fine-tuned with HyPe have lower hidden representation similarity compared to vanilla fine-tuned PLMs on the top layers. For the lower layers,
% the difference is not clear and 
three lines are entangled, and this finding is consistent with that in Section \ref{section:gen}. 

It is worth noticing that for token similarity on CoLA, although HyPe-U has lower similarity on the last layer, while has lower performance than HyPe-N in Table \ref{table-main}. There may seem a contradiction between results. However, HyPe-N achieves better similarity on other higher layers. As HyPe is added to all different layers and information from intermediate layers influences that from the last layer, the results are also consistent.

In summary, inspired by previous research on interpreting PLMs, we empirically provide an insight that HyPe may improve fine-tuning by making hidden representations isotropic-distributed.

\begin{figure}[t]
    \centering
    \includegraphics[width=0.3\textwidth]{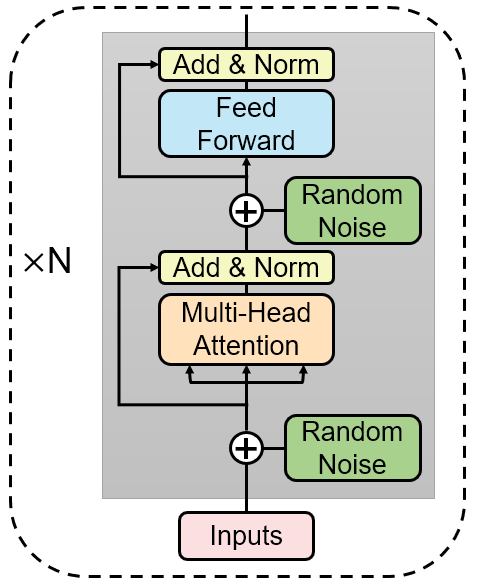}
    \caption{Besides adding noise to the hidden representations fed into Transformers layers, noise can also be added to between self-attention and feed-forward network within Transformers.}
    \label{fig:more1}
\end{figure}

\paragraph{Adding noise after self-attentions.} In HyPe, we add noise to the hidden representations between each Transformer layer, and compared to dropout, HyPe empirically shows better performance. These findings lead to this discussion of adding noise to the representations between self-attention and feed-forward network within Transformers layer like dropout, as illustrated in Figure \ref{fig:more1}. We run experiments on CoLA, STS-B, MRPC, and RTE with different schemes of adding noise.  Experiments are conducted on BERT-large.  

\begin{table*}[t]
  \centering
  \small
%   \resizebox{1.\textwidth}{!}{
  \begin{tabular}{l|cccc|c}
     & STS-B &CoLA &MRPC &RTE&AVG \\
    \hline
    HyPe&90.37&66.26&91.14&74.37&80.54 \\
    HyPe+Adding within Transformers&90.42&65.35&91.42&73.65&80.21\\
    Adding within Transformers&90.54&65.53&91.59&71.84&79.88 \\
    \end{tabular}
%   }
  \caption{Results of analysis experiments on the distribution forms and scales of the noise.}
  \label{table:more1}
\end{table*}

As shown in Table \ref{table:more1}, in terms of average scores, HyPe-N with scale $\sigma=10^{-5}$  (i.e., only adding noise between Transformers layers) shows the best performance, while adding noise only within Transformers shows the worst result among the three. When combining both positions to add noise, the performance shows no improvements on performances. 

\paragraph{Adding noise to a subset of hidden representations.} HyPe adds random noise to the hidden representations of all Transformers layers. We run further analyses by only adding noise to hidden representations fed into a subset of layers. We add normal noise with scale $\sigma=10^{-5}$ to the hidden representations in the higher 6/12 layers and lower 6/12 layers of BERT-large. The higher layers mean the layers near the classifier head, while the lower layers mean the layers near the token embedding layer. As shown in Table \ref{table:highlowlayer}, from the average scores across MRPC, STS-B, CoLA, and RTE datasets, we can conclude that 1) when adding noise on the higher layers is better than adding on the lower layers; 2) Noise added to more layers will obtain better performance. 

\begin{table*}[t]
  \centering
  \small
%   \resizebox{1.\textwidth}{!}{
  \begin{tabular}{lccccc}
    & STS-B &CoLA &MRPC &RTE&AVG \\
    \hline
    Vanilla&90.07&63.63& 90.67&72.24&79.31\vspace{0.2cm} \\
    \multicolumn{5}{l}{\textbf{HyPe on lower layers}} \\

    Lower 6 Layers&90.57&62.76&91.16&73.65&79.54 \\

    Lower 12 Layers&90.20&65.04&91.63&72.80&79.92\vspace{0.2cm}\\

    \multicolumn{5}{l}{\textbf{HyPe on higher layers}} \\

    Higher 6 Layers&90.25&64.37&91.36&73.65&79.90 \\

    Higher 12 Layers&90.27&64.36&91.53&74.73&80.22\vspace{0.2cm}\\

    HyPe & 90.37&66.26&91.14&74.37&80.54 \\
    \end{tabular}
%   }
  \caption{HyPe noise added to hidden representations of different subsets of layers.}
  \label{table:highlowlayer}
\end{table*}

\end{document}